# Identifying Symptoms of Delirium from Clinical Narratives Using Natural Language Processing

Aokun Chen, PhD[†1], Daniel Paredes, MS[†1], Zehao Yu, MS[1], Xiwei Lou, MS[1], Roberta Brunson, RN, BSN, CCRP[2], Jamie N. Thomas, RN[2], Kimberly A. Martinez, MSN, RN, CCRN[2], Robert J. Lucero, PhD, MPH, RN, FAAN[3], Tanja Magoc, PhD[4], Laurence M. Solberg, MD[5], Urszula A. Snigurska, BSN, RN[6], Sarah E. Ser, MS[7], Mattia Prosperi, PhD[7], Jiang Bian, PhD[1], Ragnhildur I. Bjarnadottir, PhD[2*], Yonghui Wu, PhD[1*]

[1]Department of Health Outcomes and Biomedical Informatics, College of Medicine, University of Florida, Gainesville, FL, USA; [2]UF Health Shands Hospital, Gainesville, FL, USA; [3]School of Nursing, University of California Los Angeles, Los Angeles, CA, USA; [4]UF Clinical and Translational Science Institute, University of Florida, Gainesville, FL, USA; [5]North Florida/South Georgia Veterans Health Service, Geriatrics Research, Education, and Clinical Center (GRECC), Gainesville, FL, USA; [6]College of Nursing, University of Florida, Gainesville, FL, USA; [7]Department of Epidemiology, College of Public Health and Health Professions & College of Medicine, University of Florida, Gainesville, FL, USA.

**Abstract**

*Delirium is an acute decline or fluctuation in attention, awareness, or other cognitive function that can lead to serious adverse outcomes. Despite the severe outcomes, delirium is frequently unrecognized and uncoded in patients' electronic health records (EHRs) due to its transient and diverse nature. Natural language processing (NLP), a key technology that extracts medical concepts from clinical narratives, has shown great potential in studies of delirium outcomes and symptoms. To assist in the diagnosis and phenotyping of delirium, we formed an expert panel to categorize diverse delirium symptoms, composed annotation guidelines, created a delirium corpus with diverse delirium symptoms, and developed NLP methods to extract delirium symptoms from clinical notes. We compared 5 state-of-the-art transformer models including 2 models (BERT and RoBERTa) from the general domain and 3 models (BERT_MIMIC, RoBERTa_MIMIC, and GatorTron) from the clinical domain. GatorTron achieved the best strict and lenient F1 scores of 0.8055 and 0.8759, respectively. We conducted an error analysis to identify challenges in annotating delirium symptoms and developing NLP systems. To the best of our knowledge, this is the first large language model-based delirium symptom extraction system. Our study lays the foundation for the future development of computable phenotypes and diagnosis methods for delirium.*

**Introduction**

Delirium, defined as any episode of acute fluctuation in attention, awareness, and other cognitive function, is one of the most common and costly iatrogenic conditions, particularly among older adults.[1–5] Among hospitalized older patients, 11-15% were estimated to have prevalent delirium while another 29-31% would develop incident delirium during the hospital stay.[6,7] Delirium is associated with serious adverse outcomes, including a higher likelihood of hospital mortality, increased length of hospital stays, greater risk of one-year mortality after discharge, functional decline, and increased caregiver burden.[2–4] However, despite its severe sequelae, delirium is frequently missed and under-coded.[8–12] In fact, one study found that less than 3% of patients with delirium had the International Classification of Diseases, Ninth Revision, (ICD-9) codes for delirium in their electronic health record (EHR) charts.[12] This was partially due to the transient nature of delirium and the difficulty in its identification.[13] As real-time identification is difficult, the prognostic study of delirium and its interventions become extremely important. Nevertheless, training a computational model to recognize delirium using only patients with certain diagnostic codes, as opposed to all patients who have delirium, is jeopardized by the error from under-coded cases. Thus, there is an urgent need for developing and deploying natural language processing (NLP) systems to identify symptoms of delirium to help identify delirium cases and inform clinicians of the interventions' effectiveness that ultimately improve the diagnosis methods for delirium.

Identifying symptoms of delirium is a typical clinical concept extraction task, which is a fundamental NLP task to identify concepts of important clinical meaning from clinical narratives. Previously, both rule-based and machine learning-based approaches have been applied. Rule-based solutions are good at recognizing concepts with fixed patterns and fewer variations such as dates and numeric sizes (e.g., tumor size), while machine learning models have good generalizability to recognize free-text concepts with diverse documentation variations such as symptoms. Clinical NLP systems such as cTAKES[14], MetaMap[15,16], and MedTagger[17] have been developed to extract general clinical concepts defined by the Unified Medical Language System (UMLS). CLAMP is a machine learning-based NLP system to facilitate the full life cycle of clinical NLP development from annotation to model training. Recent progress in NLP has greatly improved clinical concept extraction from clinical narratives. Deep learning-based large language models (LLM) trained using transformer architecture have become the state-of-the-art solution for many NLP tasks, including named-entity recognition, relation extraction, natural language inference, and question answering. Bidirectional Encoder Representations from Transformers (BERT) is one of the popular transformer structures.[18] In transformer-based models, the training process was split into pre-training with large, unlabeled data, and fine-tuning, where a small set of labeled task-specific data was involved. This structure enables the transfer learning ability: - one transformer-based model can be applied to many NLP tasks.

†Equally contributed first author
*Corresponding author, rib@ufl.edu, yonghui.wu@ufl,edu

Previous studies have applied NLP in studies of delirium diagnosis and outcomes. Wang *et al.* developed an NLP system to detect the sentiment from radiology reports to help identify delirium cases and reported that NLP-derived sentiment information improved the identification rate of delirium cases.[19] Fu *et al.* built two rule-based NLP systems, i.e., NLP-CAM and NLP-mCAM, based on the Confusion Assessment Method (CAM) and the modified Confusion Assessment Method (mCAM), to identify delirium patients from electronic health records.[20] Later, Pagali *et al.* further utilized the NLP-CAM to assess the analysis of delirium within COVID-19 patients.[13] Ge *et al.* explored SVM, CNN-LSTM, and BERT in identifying delirium-related sentences from clinical narratives.[21] They used a set of keywords to label delirium-related sentences using regular expressions. Up until now, there is no NLP solution to systematically identify and categorize delirium symptoms from clinical narratives.

To assist in and improve the diagnosis of delirium, we composed annotation guidelines, created a corpus, and developed NLP methods to extract delirium symptoms from clinical narratives. We systematically explored 5 transformer-based large language models, i.e., BERT,[18] BERT_MIMIC, RoBERTa,[22] RoBERTa_MIMIC, and GatorTron[23]. To the best of our knowledge, this is the first LLM-based delirium symptom extraction NLP system. Our study lays the foundation for the future development of computable phenotypes and diagnosis methods for delirium.

**Methods**

**Data Source**

We extracted patient EHR data from the University of Florida Health Integrated Data Repository (UF Health IDR) including UF Health Shands in Gainesville and UF Health Jacksonville. The IDR is a clinical data warehouse that collects and aggregates information from various clinical and administrative information systems across UF Health clinical and research enterprises, including the Epic EHR system. The IDR contains more than 2 billion observational facts pertaining to more than 2 million patients. We identified a cohort of 170,868 patients using the following criteria: adults (i.e., ≥ 18 years of age) who were admitted to one of the 21 medical or surgical units at the UF Health hospitals (including both UF Health Gainesville and Jacksonville) from 2012 to 2021. This study was approved by the UF Institutional Review Board (IRB #201900208).

**Table 1.** Demographics of the cohort.

| Demographics | Total (N = 170,868) |
|---|---|
| **Age, mean (SD)** | |
| Mean (SD) | 56.8 (17.2) |
| **Sex** | |
| Female (%) | 83,415 (48.82%) |
| Male (%) | 87,449 (51.18%) |
| Unknown (%) | 4 (<0.1%) |
| **Race-Ethnicity** | |
| Non-Hispanic White (%) | 113,370 (66.35%) |
| Non-Hispanic Black (%) | 42,266 (24.74%) |
| Non-Hispanic Other (%) | 2,100 (1.23%) |
| Hispanic (%) | 7,313 (4.28%) |
| Unknown (%) | 5,819 (3.41%) |

**Table 1** summarizes the demographics of this cohort. The average age of this cohort was 56.8 years, and the cohort had slightly more male patients (male: 51.18%). The majority of patients were non-Hispanic White (66.35%), and the second largest racial-ethnic group was non-Hispanic Black (24.74%). We identified a total of 6.9 million clinical notes from this cohort.

**Annotation**

As not all notes document delirium symptoms, we applied a snowball sampling strategy to curate a set of keywords used to identify the subset of notes with delirium symptoms. Specifically, domain experts (RIB, LMS, UAS) first initiated a list of keywords of delirium symptoms, which was used to identify relevant notes. We then iteratively reviewed batches of 30 notes to identify new keywords to extend the keywords list and re-identify the delirium-related notes, until there were no new keywords identified from this iterative procedure. A total number of 45 keywords were identified using this snowball strategy. Then, we filtered the 6.9 million notes using the 45 keywords to identify notes with at least 3 delirium keywords and randomly selected 600 notes for annotation. We formed an expert panel of NLP researchers (YW, AC, DP), clinical researchers (RIB, LMS, UAS, RJL), and practicing nurses (RB, JNT, KAM) to develop annotation guidelines. The expert panel reviewed delirium signs and symptoms and categorized them into 8 categories:

(1) <u>Disturbed attention</u>: defined as reduced ability to direct, focus, shift, or sustain attention, or reduced orientation to the environment (e.g., *"Disoriented", "unable to follow directions", "confused"*). Altered mental status are also included.
(2) <u>Disturbed perception</u>: defined as reduced ability to identify, organize, and interpret sensory information. Examples include hallucinations, illusions, misinterpretations, and various delusions (e.g., *"trying to poison me", "stealing from me"*).
(3) <u>Psychomotor activity</u>: defined as problematic behaviors and restless physical activity arising from mental tension; often purposeless and unintentional, or slow physical activity arising from inhibition of mental activity. For example, *"pulling off tubes", "combative", "restless", "spitting"*.
(4) <u>Fluctuations</u>: defined as changes in symptoms during the course of the day and night; for example, *"becomes more agitated", "progressively", "increasingly"*.
(5) <u>Memory deficit</u>: defined as reduced ability to encode, store, and retrieve information when needed. For example, inability to remember recent events (e.g., taking medications) or instructions (e.g., to call for help), forgetfulness, forgetting, can't recall/can't remember, no recollection, memory loss (e.g., *"forgetfulness", "did not know why he was brought here", "short term memory loss"*).
(6) <u>Consciousness level</u>: defined as waking state (wakefulness); a condition of awareness of one's surroundings, generally coupled with an ability to communicate with others or to signal understanding of what is being communicated by others. Examples include drowsy, lethargic, obtunded, stuporous, coma, etc. (e.g., *"unable to stay awake", "unresponsive", "lethargic"*).
(7) <u>Disturbed sleep</u>: defined as disturbed sleep-wake cycles circadian state characterized by partial or total suspension of consciousness, voluntary muscle inhibition, and relative insensitivity to stimulation. For example, daytime sleepiness and nighttime agitation (e.g., *"trouble falling asleep", "poor sleep", "minimal sleep"*).
(8) <u>Disorganized thinking</u>: defined as a disrupted form or structure of thinking; manifested in disorganized speech. For example, rambling or irrelevant conversation (tangentiality), a large amount of nonessential information (circumstantiality), unclear or illogical flow of ideas (loose associations), unpredictable switching from subject to subject (derailment), mumbling, incoherent, illogical, rambling (e.g., *"unable to clearly verbalize", "fixated on", "screaming incoherent words"*).

We also used an 'Other' category to capture "*hard*" cases not covered by the annotation guidelines. The categories were created and pre-defined based on the diagnostic criteria which were described in the Diagnostic and Statistical Manual of Mental Disorders, Fifth Edition (DSM-V). Following standard NLP development practice, we recruited 3 annotators and conducted training sessions to train the annotators until a good agreement score was achieved. Three annotators manually reviewed the 546 notes to identify 8 categories of delirium symptoms. We monitored the annotation agreement and periodically collected the discrepancies from annotators and instances annotated as 'Other' and discussed them in expert panel meetings. We also improved the annotation guidelines as necessary to cover new cases and solve discrepancies as needed.

**Delirium symptom extraction using transformer models**

We approached the delirium symptom extraction as a clinical concept extraction (or named-entity recognition [NER])

task and adopted the standard beginning-inside–outside (BIO) annotation format. We applied tokenization, sentence boundary detection, and BIO format transformation using a preprocessing pipeline from a previous study.[24] Then, we applied transformer models to identify delirium symptoms. Specifically, we generated distributed representations of text using the transformer models and calculated probability scores for each BIO category using a linear layer with a softmax activation function. The cross-entropy loss was used for fine-tuning.

**Transformer-based deep learning models**

We explored 5 pretrained LLMs, including 2 transformer models (BERT and RoBERTa) for general English domain and 3 transformer models (BERT_MIMIC, RoBERTa_MIMIC, and GatorTron) for clinical domain.

- **BERT and BERT_MIMIC**: Following the introduction of the transformer model by Vaswani *et al.*[25], Devlin *et al.* improved it with Bidirectional Encoder Representations from Transformers (BERT)[18]. BERT used bidirectional representations and an encoder structure that improved the performance of fine-tuning the pre-trained model. BERT_MIMIC followed the same structure of BERT but pre-trained with clinical notes from the Medical Information Mart for Intensive Care (MIMIC) dataset. In our experiment, we adopted the BERT model implemented in Hugginface.

- **RoBERTa and RoBERTa_MIMIC**: Liu *et al.* optimized the training strategies of BERT and created RoBERTa.[22] RoBERTa introduced new strategies including dynamic masking, full sentence sampling, large mini-batches, large byte level encoding, and removed next sentence prediction loss. RoBERTa_MIMIC utilized the same optimization of RoBERTa but trained over the MIMIC dataset. In our experiment, we explored the RoBERTa model implemented in Hugginface.

- **GatorTron model**: We adopted the GatorTron model in the delirium symptom extraction system. GatorTron is a BERT-style large clinical language model. We pretrained GatorTron with >90 billion words of text, including >80 billion words from >290 million notes identified at the UF Health system covering patient records from 2011–2021 from over 126 clinical departments and ~50 million encounters. These clinical narratives covered healthcare settings including but not limited to inpatient, outpatient, and emergency department visits. We used the GatorTron model with 345 million parameters for this study.**[23]**

**Training strategies**

For delirium symptom extraction and classification, we adopted the standard NER training procedure to recognize and classify delirium symptoms using a training set and a development set. We trained the models using the training set (train) of 381 notes and monitored the performance with the 55 notes as the development set (dev). The best of each transformer model was selected based on the validation performance on the development set.

**Experiment and evaluation**

We reused the pretrained models from the public GitHub repository for two transformer models from the general domain, including BERT and RoBERTa. For the two clinical transformer models, we adopted the BERT_MIMIC and RoBERTa_MIMIC models developed by fine-tuning the general models using clinical text from the MIMIC III database in our previous study.[24] The GatorTron model was developed by training from scratch using >90 billion words of text (including >82 billion words of de-identified clinical text from UF Health) in our previous studies.**[23]** We evaluated our delirium symptom extraction system with the test set of 110 notes on both strict (i.e., exact boundary surface string match and entity type) and lenient (i.e., partial boundary match over the surface string) precision, recall, and F1-score. The evaluation scores were calculated with the evaluation script from a pipeline implemented in a previous study.[24]

**Results**

After annotation, we excluded 54 notes that either were duplicated or without valid delirium mentions. We annotated a total of 2,496 concepts for 8 categories of delirium symptoms from a total of 546 clinical notes. The annotation agreement measured by F1-score among the 3 annotators improved from 29.2% to 97.1% among the different batches. All the notes were annotated independently by at least two annotators. **Table 2** shows the summary of statistics for this dataset.

**Table 2.** Summary of delirium symptom concepts.

| Symptom concepts | Train (%) | Dev (%) | Test (%) |
|---|---|---|---|
| Disturbed attention | 423 (23.51%) | 53 (23.25%) | 107 (22.81%) |
| Disturbed perception | 64 (3.56%) | 1 (0.44%) | 14 (2.99%) |
| Psychomotor activity | 819 (45.53%) | 91 (39.91%) | 205 (43.71%) |
| Fluctuations | 124 (6.89%) | 14 (6.14%) | 31 (6.61%) |
| Memory deficit | 71 (3.95%) | 2 (0.88%) | 25 (5.33%) |
| Consciousness level | 227 (12.62%) | 55 (24.12%) | 70 (14.93%) |
| Disturbed sleep | 31 (1.72%) | 5 (2.19%) | 10 (2.13%) |
| Disorganized thinking | 40 (2.22%) | 7 (3.07%) | 7 (1.49%) |

The top 3 most frequently mentioned categories of symptom concepts are psychomotor activity (train vs. dev vs. test = 45.53% : 39.91% : 43.71%), attention disturbed (23.51% : 23.25% : 22.81%), and consciousness level (12.62% : 24.12% : 14.93%). Comparing train, dev, and test set, training set had slightly more attention disturbed concepts (23.51% : 23.25% : 22.81%), psychomotor activity concepts (45.53% : 39.91% : 43.71%), and fluctuation concepts (6.89% : 6.14% : 6.61%) than the dev or test set, while also having noticeably more disturbed perception concepts (3.56% : 0.44% : 2.99%), especially compared to the dev set, while dev set had significantly more consciousness level mentions (12.62% : 24.12% : 14.93%), disorganized thinking mentions (2.22% : 3.07% : 1.49%), and slightly more disturbed sleep mentions (1.72% : 2.19% : 2.13%) than train and test set. Finally, the test set contained significantly more memory deficit concepts (3.95% : 0.88% : 5.33%) than the train and dev sets.

Table 3 compares the performance of 5 transformer models for delirium symptom extraction. GatorTron achieved the best strict and lenient F1-scores of 0.8055 and 0.8759, second by BERT_MIMIC (strict: 0.7901, lenient: 0.8724). The RoBERTa model achieved the best recall. All clinical transformer models, except RoBERTA_MIMIC, outperformed general transformer models on strict F1-score.

**Table 3.** Comparison of 5 transformer models using the overall performance.

| Model | Strict | | | Lenient | | |
|---|---|---|---|---|---|---|
| | Precision | Recall | F1-score | Precision | Recall | F1-score |
| BERT | 0.7520 | 0.8022 | 0.7763 | 0.8453 | 0.8883 | 0.8663 |
| BERT_MIMIC | 0.7705 | 0.8107 | 0.7901 | 0.8537 | 0.8920 | 0.8724 |
| RoBERTa | 0.7508 | **0.8265** | 0.7868 | 0.8265 | **0.9078** | 0.8626 |
| RoBERTa_MIMIC | 0.7728 | 0.8010 | 0.7867 | 0.8590 | 0.8799 | 0.8693 |
| GatorTron | **0.7993** | 0.8119 | **0.8055** | **0.8696** | 0.8823 | **0.8759** |

*Best Strict and Lenient precision, recall, and F1-scores are highlighted in bold.

Table 4 further breaks down the performance of GatorTron on different delirium symptoms. GatorTron achieved best extraction performance on a consciousness level and disturbed sleep with a strict F1-score over 0.9. For most of the symptoms, GatorTron demonstrated good performance with strict F1-scores between 0.73 and 0.88.

**Table 4.** Detailed performance of GatorTron for each category of delirium symptoms.

| Symptoms | Strict | | | Lenient | | |
|---|---|---|---|---|---|---|
| | Precision | Recall | F1-score | Precision | Recall | F1-score |
| Disturbed attention | 0.7960 | 0.8556 | 0.8247 | 0.8657 | 0.9305 | 0.8969 |
| Disturbed perception | 0.7200 | 0.7826 | 0.7500 | 0.7600 | 0.8261 | 0.7917 |
| Psychomotor activity | 0.8191 | 0.7918 | 0.8052 | 0.9040 | 0.8715 | 0.8874 |
| Fluctuations | 0.7755 | 0.7037 | 0.7379 | 0.8367 | 0.7593 | 0.7961 |
| Memory deficit | 0.8293 | 0.9189 | 0.8718 | 0.8537 | 0.9459 | 0.8974 |
| Consciousness level | 0.9286 | 0.8835 | 0.9055 | 0.9694 | 0.9223 | 0.9453 |
| Disturbed sleep | 0.9333 | 0.8750 | 0.9032 | 0.9333 | 0.8750 | 0.9032 |
| Disorganized thinking | 0.2857 | 0.4000 | 0.3333 | 0.4762 | 0.6667 | 0.5556 |

**Table 5.** The best performance and model for each delirium category.

| Symptoms | Model | Precision | Recall | F1-score |
|---|---|---|---|---|
| Disturbed attention | GatorTron | 0.7960 | 0.8556 | 0.8247 |
| Disturbed perception | BERT MIMIC | 0.7826 | 0.7826 | 0.7826 |
| Psychomotor activity | GatorTron | 0.8191 | 0.7918 | 0.8052 |
| Fluctuations | BERT | 0.7593 | 0.7593 | 0.7593 |
| Memory deficit | RoBERTa | 0.8537 | 0.9459 | 0.8974 |
| Consciousness level | GatorTron | 0.9286 | 0.8835 | 0.9055 |
| Disturbed sleep | GatorTron | 0.9333 | 0.8750 | 0.9032 |
| Disorganized thinking | RoBERTa | 0.3500 | 0.4667 | 0.4000 |

**Table 5** shows the strict precision, recall, and F1-score of the best performed transformer model for each delirium symptom. The best model determined by the micro-average F1-score over all categories, GatorTron, achieved the best performance in 4 out of 8 delirium symptoms (i.e., disturbed attention, psychomotor activity, consciousness level, disturbed sleep), followed by RoBERTa on 2 categories of symptoms (memory deficit, disorganized thinking), and BERT and BERT MIMIC on the remaining 2 categories of symptoms (fluctuations, disturbed perception).

**Discussion and Conclusions**

Identifying and categorizing delirium symptoms from clinical narratives is critical for the diagnosis and phenotyping of delirium. This study examined the documentation of delirium symptoms in clinical narratives and categorized them into 8 groups, created a corpus of delirium symptoms, and examined 5 transformer-based NLP models for extraction of delirium symptoms from clinical narratives. The best NLP model, GatorTron, achieved the best strict and lenient F1-scores of 0.8055 and 0.8759, respectively. All clinical transformers, except RoBERTa_MIMIC, outperformed the general-purpose transformers trained using general English text by a large margin, which is consistent with our previous observation that domain-specific transformers were outperforming general purpose transformer models.

Among the 8 categories of delirium symptoms, all transformers performed well for identifying memory deficit, consciousness level, and disturbed sleep symptoms, but struggled for disorganized thinking symptoms. We conducted an error analysis and generated a confusion matrix based on the best model, i.e., GatorTron. **Figure 1** shows the confusion matrix, where the true labels are presented on the y-axis and the predicted labels are on the x-axis. As shown in **Figure 1**, most errors in identifying "disorganized thinking" came from false negatives – our system missed the concepts labeled by human experts. Among the annotated delirium symptoms, "disorganized thinking" encompasses complicated situations, i.e., irrelevant conversation (tangentiality), a large amount of nonessential information (circumstantiality), unclear or illogical flow of ideas (loose associations), unpredictable switching from subject to subject (derailment), mumbling, incoherent, illogical, rambling. When combined with the lack of mentions in the training set (2.22% of all concepts), the impact on the performance of our clinical NLP system was much higher than the symptoms with simpler situations, e.g., "disturbed sleep" and "fluctuation".

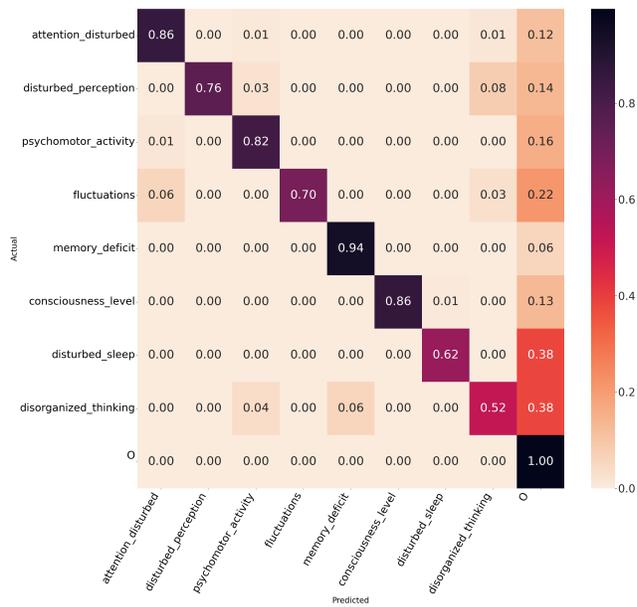

**Figure 1.** Confusion matrix *of the GatorTron model,* where the true labels are presented on the y-axis and the predicted labels are on the x-axis.

Delirium is a complex condition that is difficult to diagnose, which is reflected in the annotation of our corpus.[26,27] The judgement of whether a phrase is a delirium symptom or not highly depends on the context. For example, certain behaviors, such as '*pacing'*, '*restlessness'*, or '*repeatedly asking questions*', could be symptoms of an altered mental state, or simply be a normal reaction to stress related to a healthcare issue. In addition, some well-known serious potential symptoms of delirium, such as hallucinations or delusions, were rarely documented in the clinical notes. Instead, clinicians often describe details of occurrences where patients were behaving erratically, often using direct quotes of irrational statements of patients. Therefore, our annotators annotated many long phrases (e.g., "*claimed nurses were trying to kill him*" – '*disturbed perception'*), in which they found it was difficult to be consistent on the boundaries. This was reflected by the gap between strict and lenient F1 scores. Another unique difficulty related to this annotation task was the importance of patients' intent versus ability. For example, documentation of a patient being "*unable to follow commands*" was consistently annotated as a delirium symptom in many instances. However, a higher level of uncertainty emerged when a patient was described as "*unwilling to follow commands*" or "*did not follow commands*". In those instances, more context from the document-level would be required to determine whether this was a potential symptom of delirium or not. There are also challenges from differentiating a clinical judgement of delirium symptoms and potential/historical risks. For example, clinical notes often include documentation of a history of symptoms or an assessment that the patient is at risk of or if certain behaviors or conditions met, e.g., "*Pt has bilateral wrist restraints <u>for risk of pulling out medical equipment</u>*." and "*Patient <u>attempted to remove trach, DHT, and JP drain</u>*".

Compared with previous NLP studies on delirium patients, our study categorized detailed delirium symptoms into 8 categories, which capture more detailed information about the progression of the disease; we also applied state-of-the-art clinical transformer models to ensure the accuracy of extraction. Our future work will explore the NLP-extracted delirium symptoms for computable phenotyping of delirium to help develop efficient diagnosis methods.

**Acknowledgement**

This study was partially supported by grants from the National Institutes of Health (NIH), National Institute on Aging, R33AG062884, R56AG069880, R01AG080624, R01AG080991, R01AG076234, a Patient-Centered Outcomes Research Institute® (PCORI®) Award (ME-2018C3-14754), and the UF Clinical and Translational Science Institute (UL1TR001427). The content is solely the responsibility of the authors and does not necessarily represent the official views of the funding institutions. We would like to thank our practicing nurse annotators (RB, JNT, KAM) for their valuable contributions to this study despite the challenges posed by the COVID-19 pandemic. We gratefully acknowledge the support of NVIDIA Corporation and NVAITC with the donation of the GPUs used for this research.